\pdfoutput=1
\documentclass[11pt]{article}

\usepackage{acl}

\usepackage{times}
\usepackage{latexsym}
\usepackage[T1]{fontenc}
\usepackage[utf8]{inputenc}
\usepackage{microtype}
\usepackage{inconsolata}
\usepackage{graphicx}
\usepackage{booktabs}
\usepackage{amsmath}
\usepackage{amssymb}
\usepackage{multirow}
\usepackage{xcolor}
\usepackage{url}
\usepackage{enumitem}

\newcommand{\ActorEightBBase}{{0.862}}
\newcommand{\ActorEightBCI}{{[-0.009, +0.050]}}
\newcommand{\ActorEightBDelta}{{+0.020}}

\newcommand{\ActorFourBBase}{{0.652}}
\newcommand{\ActorFourBCI}{{[+0.039, +0.183]}}
\newcommand{\ActorFourBDelta}{{+0.110}}

\newcommand{\ActorFourteenBBase}{{0.836}}
\newcommand{\ActorFourteenBCI}{{[-0.003, +0.028]}}
\newcommand{\ActorFourteenBDelta}{{+0.012}}

\newcommand{\ActorScales}{{4B, 8B and 14B}}
\newcommand{\ActorWallHours}{{43}}

\newcommand{\AurcCI}{{[-0.0633, -0.0064]}}
\newcommand{\AurcDelta}{{-0.0362}}
\newcommand{\AurcExcl}{{excludes}}

\newcommand{\BestDimAUC}{{0.753}}
\newcommand{\BestDimName}{{specificity}}
\newcommand{\BigDiscCI}{{[+0.061, +0.149]}}
\newcommand{\BigDiscDelta}{{+0.103}}

\newcommand{\BigJudgeAlone}{{0.728}}
\newcommand{\BigJudgeBTwoCI}{{[+0.002, +0.038]}}
\newcommand{\BigJudgeBTwoDelta}{{+0.020}}
\newcommand{\BigJudgeBTwoRefitCI}{{[-0.001, +0.048]}}
\newcommand{\BigJudgeBZeroCI}{{[+0.094, +0.208]}}
\newcommand{\BigJudgeBZeroDelta}{{+0.152}}
\newcommand{\BigJudgeCI}{{[+0.020, +0.081]}}
\newcommand{\BigJudgeDelta}{{+0.050}}
\newcommand{\BigJudgeName}{{Qwen3~32B}}
\newcommand{\BigJudgeRefitCI}{{[+0.013, +0.095]}}
\newcommand{\BigJudgeRefitExcl}{{excludes}}

\newcommand{\BigRedactMatchedCI}{{[-0.030, +0.079]}}
\newcommand{\BigRedactMatchedDelta}{{+0.023}}

\newcommand{\BlockConfirmCI}{{[+0.029, +0.095]}}
\newcommand{\BlockConfirmDelta}{{+0.061}}
\newcommand{\BlockConfirmN}{{943}}
\newcommand{\BlockPilotCI}{{[-0.004, +0.066]}}
\newcommand{\BlockPilotDelta}{{+0.031}}
\newcommand{\BlockPilotN}{{312}}

\newcommand{\CondTerminalAUCalone}{{0.831}}
\newcommand{\CondTerminalCI}{{[+0.027, +0.087]}}
\newcommand{\CondTerminalDelta}{{+0.056}}
\newcommand{\ControlAUCalone}{{0.728}}
\newcommand{\ControlDeltaAUC}{{+0.053}}
\newcommand{\ControlDetectRate}{{100\%}}
\newcommand{\ControlDraws}{{15}}

\newcommand{\CostFullBaseAUC}{{0.826}}

\newcommand{\CostNoWallBaseAUC}{{0.826}}
\newcommand{\CostNoWallCI}{{[+0.019, +0.081]}}
\newcommand{\CostNoWallDelta}{{+0.049}}
\newcommand{\CostTokensOnlyBaseAUC}{{0.827}}
\newcommand{\CostTokensOnlyCI}{{[+0.019, +0.082]}}
\newcommand{\CostTokensOnlyDelta}{{+0.050}}

\newcommand{\ErateEThree}{{16.6\%}}

\newcommand{\ErateGivenFailing}{{28.8\%}}

\newcommand{\EthreeRemoved}{{5}}
\newcommand{\EvidenceJudge}{{Qwen3-32B}}

\newcommand{\FlexCIAUC}{{[+0.011, +0.080]}}
\newcommand{\FlexDeltaAUC}{{+0.046}}

\newcommand{\FourteenBCI}{{[-0.000, +0.049]}}
\newcommand{\FourteenBDelta}{{+0.023}}

\newcommand{\GoldOverlapAUCalone}{{0.536}}
\newcommand{\GoldOverlapCI}{{[-0.016, +0.034]}}
\newcommand{\GoldOverlapDelta}{{+0.007}}
\newcommand{\GridBOneFree}{{+0.055}}

\newcommand{\GridBOneOverall}{{+0.030}}

\newcommand{\JudgeCompletionMTok}{{2.79}}
\newcommand{\JudgePromptMTok}{{42.5}}
\newcommand{\LadderGapAllExcl}{{all four}}
\newcommand{\LadderGapMax}{{+0.036}}
\newcommand{\LadderGapMaxCI}{{[+0.011, +0.062]}}
\newcommand{\LadderGapMin}{{+0.027}}
\newcommand{\LadderGapMinCI}{{[+0.008, +0.046]}}

\newcommand{\LadderGapN}{{4}}
\newcommand{\LadderGapPMax}{{0.010}}
\newcommand{\LadderGapThinkCI}{{[+0.007, +0.054]}}
\newcommand{\LadderGapThinkDelta}{{+0.031}}
\newcommand{\LeakCIAUC}{{[-0.012, +0.043]}}
\newcommand{\LeakCorr}{{0.87}}
\newcommand{\LeakDeltaAUC}{{+0.016}}
\newcommand{\LeakExcl}{{includes}}
\newcommand{\LeakPval}{{0.260}}
\newcommand{\LoroNfolds}{{5}}
\newcommand{\LoroNpositive}{{4}}

\newcommand{\MainExcl}{{excludes}}

\newcommand{\MainUpperBound}{{+0.081}}

\newcommand{\MeanFiveAUCalone}{{0.728}}

\newcommand{\MidRedactMatchedCI}{{[-0.089, -0.005]}}
\newcommand{\MidRedactMatchedDelta}{{-0.047}}

\newcommand{\MidRedactP}{{0.03}}
\newcommand{\MinDetectableAUC}{{0.650}}
\newcommand{\MinDetectableDelta}{{+0.024}}

\newcommand{\Nactors}{{3}}
\newcommand{\Nepisodes}{{1,264}}
\newcommand{\NexecFail}{{9}}
\newcommand{\Npilot}{{316}}

\newcommand{\NposEThree}{{208}}

\newcommand{\Nreplays}{{4,621}}
\newcommand{\Nrepos}{{7}}
\newcommand{\Nscorercalls}{{20,825}}
\newcommand{\Ntasks}{{158}}
\newcommand{\NtermExecutableFailing}{{722}}
\newcommand{\NtermExecutablePassing}{{186}}
\newcommand{\NtermNoArtifact}{{115}}
\newcommand{\NtermNotExecutable}{{232}}

\newcommand{\Nunits}{{1,255}}
\newcommand{\OverallAUCalone}{{0.705}}
\newcommand{\OverallVsMeanGap}{{0.023}}
\newcommand{\PairCI}{{[0.58, 0.86]}}
\newcommand{\PairConcordant}{{26}}
\newcommand{\PairDecided}{{35}}
\newcommand{\PairDiscordant}{{9}}
\newcommand{\PairN}{{46}}
\newcommand{\PairP}{{0.006}}
\newcommand{\PairRate}{{74.3\%}}
\newcommand{\PairRateTiesSplit}{{68.5\%}}
\newcommand{\PairTies}{{11}}
\newcommand{\PairVerdict}{{better than chance}}
\newcommand{\ParseRate}{{99.2\%}}
\newcommand{\PermNullPninetyfive}{{+0.008}}
\newcommand{\PermPval}{{0.000}}

\newcommand{\RedactInterCI}{{[+0.010, +0.126]}}
\newcommand{\RedactInterDelta}{{+0.069}}
\newcommand{\RedactInterExcl}{{excludes}}
\newcommand{\RedactInterN}{{1151}}
\newcommand{\RedactInterP}{{0.02}}

\newcommand{\RefitBestDimCI}{{[+0.014, +0.116]}}
\newcommand{\RefitBestDimDelta}{{+0.060}}
\newcommand{\RefitBestDimExcl}{{excludes}}
\newcommand{\RefitBestDimName}{{specificity}}
\newcommand{\RefitBestDimP}{{0.00}}
\newcommand{\RefitContamCI}{{[-0.005, +0.027]}}
\newcommand{\RefitContamExcl}{{includes}}

\newcommand{\ReplayStablePct}{{100.0\%}}
\newcommand{\RepoList}{{\texttt{sympy}, \texttt{sphinx}, \texttt{xarray}, \texttt{pytest}, \texttt{requests}, \texttt{pylint}, \texttt{seaborn}}}
\newcommand{\RolloutsEightB}{{4}}
\newcommand{\RolloutsFourB}{{2}}

\newcommand{\ScorerAvgAUC}{{0.655}}
\newcommand{\ScorerAvgCI}{{[+0.001, +0.044]}}
\newcommand{\ScorerAvgDelta}{{+0.021}}
\newcommand{\ScorerICC}{{0.88}}

\newcommand{\ScorerScales}{{4B, 8B, 14B and 32B}}
\newcommand{\ScorerSeedCorr}{{0.89}}

\newcommand{\SmallJudgeMaxAlone}{{0.700}}
\newcommand{\SmallJudgeRange}{{4B to 14B}}

\newcommand{\StratCIAUC}{{[+0.044, +0.169]}}
\newcommand{\StratDeltaAUC}{{+0.103}}
\newcommand{\StratExcl}{{excludes}}
\newcommand{\StratN}{{722}}
\newcommand{\StratNpos}{{208}}

\newcommand{\StratOutN}{{533}}
\newcommand{\StratRateDeficient}{{17.2\%}}
\newcommand{\StratRateSupported}{{42.0\%}}
\newcommand{\StratRefitCI}{{[+0.022, +0.209]}}
\newcommand{\StratRefitExcl}{{excludes}}

\newcommand{\TermLeakCIAUC}{{[+0.061, +0.149]}}
\newcommand{\TermLeakDeltaAUC}{{+0.103}}
\newcommand{\ThinkAlone}{{0.700}}
\newcommand{\ThinkCI}{{[-0.003, +0.041]}}
\newcommand{\ThinkDelta}{{+0.018}}
\newcommand{\ThinkDiscCI}{{[+0.024, +0.119]}}
\newcommand{\ThinkDiscDelta}{{+0.071}}
\newcommand{\ThinkN}{{1,220}}
\newcommand{\TopRepoPosShare}{{91\%}}
\newcommand{\TopRepoShare}{{70\%}}
\newcommand{\TrajCharCap}{{24{,}000}}
\newcommand{\TruncatedPct}{{0.2\%}}

\newcommand{\WorstDimAUC}{{0.517}}
\newcommand{\WorstDimName}{{recovery}}

\title{Endpoint-Conditioned Process Value: What an Endpoint-Blind Process Score
Adds Once Terminal Status and Cost Are Known}

\author{
  \textbf{Chris D Yan\textsuperscript{*}},
  \textbf{Ken Wang\textsuperscript{*}},
  \textbf{MarkX Qian\textsuperscript{*}},
  \textbf{Jin Jiang\textsuperscript{*}},
\\
  \textbf{Albus Wu\textsuperscript{$\dagger$}},
  \textbf{Wenhao Wang\textsuperscript{$\ddagger$}}
\\ 
   Vast Intelligence Lab, Sydney, Australia
\\
  \textsuperscript{*}Equal Contribution \quad
  \textsuperscript{$\dagger$}Project Lead \quad
  \textsuperscript{$\ddagger$}Corresponding Author
}

\begin{document}
\maketitle

\begin{abstract}
Trajectory metrics promise information beyond final-answer accuracy, but no one
has measured what a process score adds once an evaluation knows the outcome and
the cost. We build that comparison: a sealed consequence needing an artifact
neither agent nor judge ever sees, an operational baseline of terminal status
and cost, and judge scale and evidence access as experimental factors. On
\Nunits{} trajectories over \Ntasks{} test-generation tasks in \Nrepos{} Python
repositories, the answer is gated on the judge. Four judges from
\SmallJudgeRange{}, one reasoning before scoring, add nothing an
interval separates from zero; a \BigJudgeName{} judge on identical trajectories
adds \BigJudgeDelta{} AUC (95\% CI \BigJudgeCI{}) and survives a refitting
bootstrap (\BigJudgeRefitCI{}). Most of what it adds is recoverable from eight
behavioural counters, so a harness should log them (the judge adds
\BigJudgeBTwoDelta{} beyond them).
What the judge is shown moves it further: disclosing the terminal verdict
inflates its standalone discrimination by \BigDiscDelta{}, while reference
disclosure and tool-output redaction move it by nothing an interval separates
from zero. A process metric reported without the judge's
scale and evidence access does not identify a quantity. We release the
protocol, the sealed-endpoint harness, and all trajectories.
 \end{abstract}

\section{Introduction}
\label{sec:intro}

Agent evaluation is moving from final answers to trajectories. A benchmark that
reports only whether the agent solved the task is said to miss how it got there,
and a growing family of process metrics scores the trajectory itself: efficiency,
hallucination and adaptivity \citep{kim2026trace}, tool-use rationale and planning
quality \citep{wang2025mcpbench}, or the whole reasoning process judged by another
agent \citep{zhuge2024agentjudge}. The premise is that these scores carry
information the outcome does not.

That premise is rarely tested in the form that matters operationally. A team
choosing whether to add a process metric to its harness already knows two things
for free: whether the artifact passed its visible check, and what it cost to
produce. The useful question is not whether the process score correlates with
quality. It is whether the score still predicts a consequence the team cares about
\emph{after} conditioning on the outcome and the cost it already has. That is the
incremental-validity question \citep{sechrest1963incremental,
hunsley2003incremental}, and process metrics for agents have not been subjected to
it.

\paragraph{Position.} We measure it directly, and the answer turns on the judge.
An endpoint-blind process score does carry information about a sealed consequence
beyond the outcome and the cost, but only when the judge is large enough: four
judges from \SmallJudgeRange{}, one of them allowed to reason before scoring, add
nothing an interval separates from zero, while a \BigJudgeName{} judge on the
identical trajectories does. Even then most of what it knows is recoverable by
counting. And on both judges where we ran the contrast, what the judge is shown moves it
more than the process evidence does.

Testing this requires three design obligations. The consequence must be
\emph{sealed}, computed from executions needing an artifact the agent and the judge
never saw, so score and endpoint do not share a measurement. The baseline must be
the \emph{operational} one, terminal status plus cost. And evidence access must be
an \emph{experimental factor}, so contamination is measured rather than assumed
absent.

We build all three on repository-level regression-test generation
\citep{mundler2024swtbench}. An agent gets a real checkout of a repository with an
unfixed bug and five tools, one of which executes candidate tests against the
buggy code. What it submits is executed once at the base commit, giving the
terminal status a developer can see. Three further executions, using the sealed
gold patch and an unrelated patch from the same repository, give the endpoint: the
artifact must flip on the real fix, reproduce that verdict on an independent
replay, and keep failing under unrelated source churn.

On \Nunits{} trajectories from \Nactors{} actor scales over \Ntasks{} tasks in
\Nrepos{} repositories, the four smaller judges reach at most \SmallJudgeMaxAlone{}
AUC against the sealed endpoint alone and the best is worth \FourteenBDelta{}
(\FourteenBCI{}) over terminal status and cost, reasoning included. The
\BigJudgeName{} judge reaches \BigJudgeAlone{} alone and \BigJudgeDelta{}
(\BigJudgeCI{}) incrementally, surviving a bootstrap that refits every fold
(\BigJudgeRefitCI{}), and it beats each smaller judge by \LadderGapMin{} to
\LadderGapMax{} in a paired contrast on the same trajectories.

Our contributions are threefold:

\begin{itemize}[leftmargin=*]
\item \textbf{A protocol for endpoint-conditioned process value.}
We introduce a sealed replay endpoint that is independent of the terminal
evaluator and reproducible to \ReplayStablePct{}, together with an evidence
firewall treated as an experimental factor, out-of-fold estimation grouped by
repository, and a matched-strength synthetic control showing that the estimator
detects incremental value when it exists (\S\ref{sec:method}).

\item \textbf{The measurement: process value is gated on judge scale.}
We evaluate a ladder of five judges spanning \SmallJudgeRange{} and
\BigJudgeName{}, crossed with the information already available to the baseline
and calibrated against a matched-strength detectability floor. We replicate the
analysis across three actor scales, two endpoint definitions, and both fold
groupings (\S\ref{sec:results}), with an additional cross-family check in
Appendix~\ref{app:rebuttal}.

\item \textbf{Outcome access moves the judge; process evidence barely does.}
We show that disclosing the terminal verdict raises the primary judge's
standalone discrimination by \BigDiscDelta{} (\BigDiscCI{}) on identical
trajectories and the reasoning judge's by \ThinkDiscDelta{} (\ThinkDiscCI{}).
In contrast, disclosing the reference solution or withholding every raw tool
output has little effect: among three controlled manipulations of the judge's
evidence, only access to the outcome substantially changes its discrimination
(\S\ref{sec:analysis}).
\end{itemize}

This is \textbf{not} an argument that process supervision is worthless. Process
rewards demonstrably help when used to \emph{select or train}
\citep{lightman2024verify}. The claim is about evaluation: a process metric's worth is a property of the judge as much as of the trajectory, so a number reported without the judge's scale and its evidence access does not identify a quantity.
 \section{Related Work}
\label{sec:related}

\subsection{Trajectory-level agent evaluation} 

A growing line of work proposes
trajectory metrics alongside final-answer matching: TRACE scores efficiency,
hallucination and adaptivity from an evidence bank \citep{kim2026trace}; MCP-Bench
combines executable tool checks with LLM-judged rationale and planning rubrics
\citep{wang2025mcpbench}; Agent-as-a-Judge evaluates the reasoning trajectory
rather than the artifact \citep{zhuge2024agentjudge}. These establish that process
scores are \emph{measurable} and disagree with final accuracy. None establishes
what the disagreement is worth once terminal status and cost are known. That gap is
what we measure.

\subsection{Repository-level test generation}

SWT-Bench turns SWE-bench repair
tasks into test-generation tasks, scoring an artifact by whether it fails before
the gold patch and passes after \citep{mundler2024swtbench, jimenez2024swebench,
yang2024sweagent}. We adopt that substrate and tighten the endpoint: a test that
flips because the source changed at all, or that flips only once, is not a
regression test a maintainer keeps, so we also require replay stability and
specificity against an unrelated patch.

\subsection{Surrogates and incremental validity} 

Clinical statistics has a long
tradition of asking whether a cheap intermediate measurement can stand in for the
outcome that matters, and a longer one of finding that it cannot
\citep{prentice1989surrogate, fleming1996surrogate}. Psychometrics formalised the
matching question for predictors: a new measure earns its place only by improving
prediction over those already available \citep{sechrest1963incremental,
hunsley2003incremental}. Process metrics are surrogates in exactly this sense, and
we apply the corresponding test. Related concerns about optimising a proxy motivate
work on reward hacking \citep{skalse2022reward, manheim2018goodhart,
denison2024sycophancy} and on imperfect verifiers limiting the returns to
resampling \citep{stroebl2024inference}; ours is a measurement, not a mitigation.
\citet{kapoor2024agents} argue agent evaluation needs cost-aware baselines, which
is why cost is inside ours.
 \section{Measuring Incremental Process Value}
\label{sec:method}

\begin{figure*}[t]
\centering
\includegraphics[width=\textwidth]{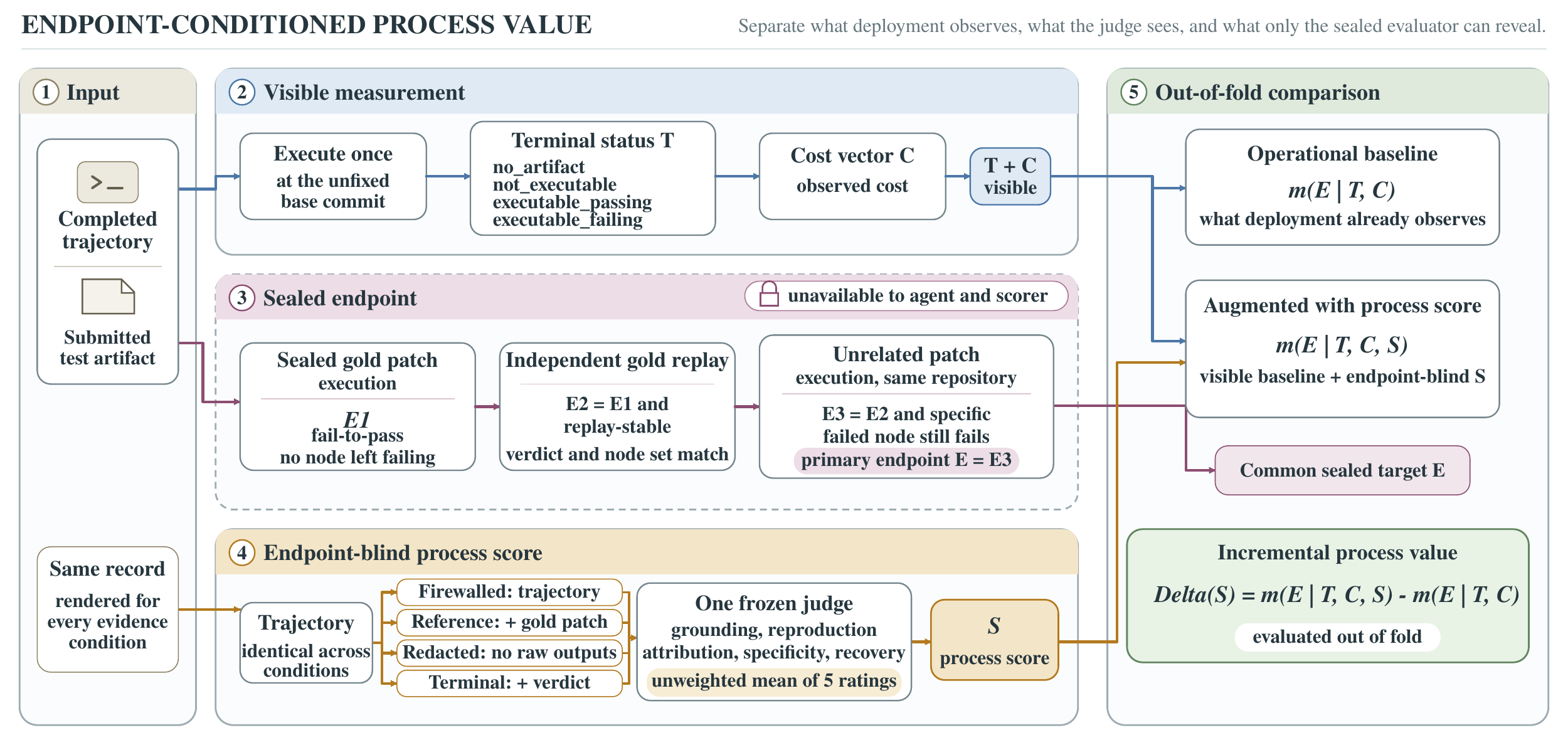}
\caption{The measurement. One execution at the base commit gives the terminal
status a developer can see; three further executions, using the sealed gold patch
and an unrelated patch, give the endpoint nobody in the scoring loop can see. A
frozen judge scores the trajectory under four evidence conditions.}
\label{fig:pipeline}
\end{figure*}

\subsection{The estimand}

As Figure~\ref{fig:pipeline} summarizes, let a completed trajectory have terminal
status $T$, cost vector $C$, and
endpoint-blind process score $S$, and let $E$ be a sealed downstream consequence,
not computable from anything the agent or the scorer observed. The quantity of
interest is the \emph{incremental value} of $S$,
\begin{equation}
\Delta(S) \;=\; m\!\left(E \mid T, C, S\right) \;-\; m\!\left(E \mid T, C\right),
\label{eq:estimand}
\end{equation}
for a proper scoring or discrimination measure $m$ evaluated out of fold. This is
the classical incremental-validity question \citep{sechrest1963incremental,
hunsley2003incremental} applied to process metrics, and it is harder to satisfy
than the correlational claim usually reported: a score can correlate with the
endpoint and still add nothing, if what it knows is already carried by the outcome
and the cost.

Three properties make $\Delta(S)$ interpretable, and all are design obligations
rather than analysis choices. The endpoint must be \textbf{sealed}: if $E$ comes
from the execution that produced $T$, then $\Delta(S)$ measures agreement between
two views of one measurement. The baseline must be the \textbf{operational} one,
since a deployment already knows the visible outcome and the cost. And the score
must be \textbf{endpoint-blind}, which makes evidence access an experimental
factor rather than an implementation detail.

\subsection{Task, artifact and terminal status}

The substrate is repository-level regression-test generation in the style of
SWT-Bench \citep{mundler2024swtbench}, on SWE-bench-Verified tasks
\citep{jimenez2024swebench}. An agent receives a checkout at a commit containing an
unfixed bug, the issue text, and five tools: \texttt{list\_dir},
\texttt{read\_file}, \texttt{search}, \texttt{run\_test} and \texttt{submit}.
\texttt{run\_test} writes a candidate test into the current checkout, executes it
and returns raw \texttt{pytest} output, so the agent can see whether its test fails
and why -- exactly what a developer has before a fix exists. It cannot see the
fix.

The submitted artifact is executed once by a frozen evaluator at the base commit.
That single execution defines the \textbf{terminal status} $T$, which takes one of
three values: \textsc{not\_executable} (nothing collected, a launch failure or a
timeout), \textsc{executable\_passing} (collects, no node fails), and
\textsc{executable\_failing} (collects and at least one node fails).
Trajectories that submit nothing are recorded as \textsc{no\_artifact}.

\subsection{The sealed endpoint}

Three further executions use material the agent and the scorer never see: the
sealed gold patch, an independent replay of it, and an unrelated patch drawn from
the same repository. They define a nested family:

\begin{itemize}
\itemsep0.15em
\item \textbf{$E_1$ fail-to-pass.} A node that failed at base passes under the gold
patch, and no node is left failing.
\item \textbf{$E_2 = E_1 \wedge$ replay-stable.} An independent re-execution
reproduces the verdict and the same flipped node set.
\item \textbf{$E_3 = E_2 \wedge$ specific.} Under an unrelated patch, a node that
failed at base still fails, so the flip under the gold patch is attributable to
the real fix rather than to any perturbation of the source.
\end{itemize}

$E_3$ is the primary endpoint. It is the property a maintainer actually wants from
a generated regression test: it must fire on this bug, keep firing under
unrelated churn, and stop firing once the bug is fixed. The nesting means $E_1$ and
$E_2$ are pre-declared relaxations, not post-hoc alternatives.

\subsection{Process score and evidence firewall}

A frozen scorer reads a rendered trajectory and rates five rubric dimensions on
$1$--$5$ (grounding, reproduction, attribution, specificity, recovery) plus its own
$0$--$100$ overall score. It is asked to judge the quality of the work and is never
asked to predict anything. The primary aggregate is the unweighted mean of the five
dimensions; the scorer's own overall field is reported separately, and the gap
between the two turns out to matter (\S\ref{sec:analysis}).

Four evidence conditions differ only in a block prepended to the identical rendered
trajectory:

\begin{itemize}
\itemsep0.15em
\item \textsc{firewalled}: the trajectory alone. Primary.
\item \textsc{reference}: plus the sealed gold patch. This is the condition created
by any harness that lets a judge see the reference solution.
\item \textsc{redacted}: raw tool outputs withheld, so only the agent's own account
of its work survives.
\item \textsc{terminal}: plus the terminal evaluator's verdict, stated in words.
\end{itemize}

Neither \textsc{reference} nor \textsc{terminal} is a straw man: both are
defaults in deployed harnesses. Comparing each against \textsc{firewalled} on the
same trajectories, scored by the same judge, separates process signal from
disclosure. The judge must be held fixed: a contrast whose arms are scored by
different judges sums an evidence effect and a judge effect, and \S\ref{sec:results}
shows the judge term can be the larger of the two.

\subsection{Estimator and inference}

We predict $E$ with $L_2$-penalised logistic regression over three nested feature
blocks: $B_0$ terminal status; $B_1 = B_0 +$ cost (log completion tokens, log
prompt tokens, log wall time, turns); and $B_2 = B_1 +$ behaviour counters (reads,
searches, test runs, listings, parse failures, whether the agent ever ran its own
test, log artifact length, distinct files read). $B_2$ is deliberately hostile to
our own hypothesis: it hands the baseline every cheap trajectory statistic that
does not require a judge.

Predictions are out of fold under two groupings. \textbf{By repository}
(leave-one-repository-out) is primary and asks whether the relationship transfers
to a repository never seen in training. \textbf{By task} (five folds) is
secondary. Metrics are AUC, log loss, Brier score and expected calibration error
over ten quantile bins \citep{guo2017calibration}.

Intervals come from a cluster bootstrap that resamples \emph{tasks}, because the
rollouts of one task are not independent \citep{efron1993bootstrap}. Condition
contrasts are paired: the same resampled index set is applied to both conditions,
which is far more able to separate them than comparing two independently computed
intervals. A permutation null is obtained by permuting the process score across
units and rerunning the entire out-of-fold procedure.
 \section{Experimental Setup}
\label{sec:setup}

\subsection{Substrate}

We use \Ntasks{} SWE-bench-Verified tasks across \Nrepos{} Python repositories
(\RepoList{}), each with a frozen base commit, a per-repository virtual
environment, a validated gold patch, and a reference test file. Every task was
verified before this study to have a gold patch that applies and an environment
that executes, so an execution failure here is a property of the generated
artifact rather than of the substrate. All task material comes from public issue
reports and source code, used under the repositories' open-source licences and
for the purpose those artifacts were published for; the study involves no human
subjects, no annotators and no personal data.

Execution is hermetic: a fresh hard reset of a git worktree, no network, fixed
\texttt{PYTHONHASHSEED}, and a per-test watchdog. A bundle producing no
\texttt{pytest} output is recorded as an executor failure and excluded rather than
scored, so machine load cannot masquerade as a treatment effect.

\subsection{Actors and rollouts}

Three open-weight actors spanning \ActorScales{} -- \texttt{Qwen3-4B},
\texttt{Qwen3-8B} and \texttt{Qwen3-14B} \citep{yang2025qwen3}, all
AWQ-quantised \citep{lin2024awq} and served with vLLM \citep{kwon2023vllm}.
Decoding is temperature $0.7$, top-$p$ $0.8$, thinking mode disabled, with a seed
derived from a stable digest of the task id and the rollout index. The action
budget is $14$. The 8B actor contributes \RolloutsEightB{} rollouts per task and
the 4B and 14B actors \RolloutsFourB{} each, giving \Nunits{} completed
trajectories over \Ntasks{} tasks. \NexecFail{} episodes exceeded the serving
context limit, returned no trajectory, and are excluded as executor failures.

Every trajectory is stored raw: each assistant message, the parsed action, the
complete tool output the environment returned, per-turn token counts and wall
time. No summary stands in for the record the scorer reads.

\subsection{Scorers}

Five judges span \ScorerScales{}, all AWQ-quantised and all served locally:
\texttt{Qwen3-4B}, \texttt{Qwen3-8B}, \texttt{Qwen3-14B}, the same 14B with
thinking mode enabled, and \texttt{Qwen3-32B}, which is the primary judge. All
decode at temperature $0$; thinking is disabled for every judge except the one
where it is the manipulated variable, which decodes with a larger token budget to
leave room for its reasoning. Every judge scores every trajectory under the
\textsc{firewalled} condition, so the ladder in \S\ref{sec:results} is
within-trajectory. All four evidence conditions were run in full on the primary
judge, so every evidence contrast holds the judge fixed and varies only what the
judge was shown. \textsc{terminal} was additionally run on the reasoning judge and
\textsc{redacted} on the 14B, so both disclosure and redaction are measured at two
scales rather than one. The rubric text, rendering
budget and decoding parameters are frozen and hashed; the four evidence conditions
differ only in the prepended block. Because temperature-$0$ decoding is
reproducible here, judge reliability is measured separately at temperature $1$
with three seeds. Rubric parsing
succeeded on \ParseRate{} of scorer calls.

\subsection{Pilot and confirmatory sets}

Two specification choices were made after inspecting a pilot of \Npilot{}
8B episodes (\BlockPilotN{} of which survive the executor-failure exclusion and
form the pilot block in Table~\ref{tab:generality}): the aggregate is the unweighted mean of the five dimensions rather
than the scorer's own overall field, and a secondary analysis is restricted to the
stratum where the endpoint is not determined by construction. Both are recorded
with dates in the frozen protocol shipped with the code, and every confirmatory
number is re-estimated on data that did not exist when they were made: the later
8B rollouts, the entire 4B actor, and every 14B scorer pass.

\subsection{Release}

The protocol with its dated amendments, the harnesses, the analysis and figure
code, and all \Nunits{} trajectories with their endpoint and scorer records are
released under a permissive licence in the supplementary material. Upstream
repositories, task metadata and model checkpoints are not redistributed; they are
public and named in Appendix~\ref{app:substrate}.

\subsection{Compute}

The campaign ran on two shared hosts, one NVIDIA A40 and one NVIDIA H100, one
model resident at a time on a single card of each: \Nepisodes{} actor episodes,
\Nreplays{} sealed-endpoint executions and \Nscorercalls{} scorer calls. Acting
consumed \ActorWallHours{} hours of wall-clock. Judging is reported in tokens
rather than hours, because the scorer records usage per call but not wall time
and both hosts were shared throughout: the scorer calls consumed
\JudgePromptMTok{}M prompt and \JudgeCompletionMTok{}M completion tokens across
all judges and conditions. We report both so that others can judge the cost of
replication \citep{kapoor2024agents}. Every model is open-weight and served
locally; no closed-weight model and no external API was used anywhere in the
study.

An independent repeat of the replay reproduces every endpoint variable exactly,
including which patch was drawn as the placebo.
 \section{Results}
\label{sec:results}

\begin{table*}[t]
\centering\small
\setlength{\tabcolsep}{3pt}
\begin{tabular}{lrrlll}
\toprule
Judge & $n$ & AUC alone & vs.\ terminal + cost & vs.\ + counters
& gap to \textsc{32b} \\
\midrule
Qwen3-4B & 1255 & 0.642 & $+0.014$ [-0.011, +0.050] & $+0.010$ [-0.017, +0.033] & $+0.036$ [+0.011, +0.062] \\
Qwen3-8B & 1255 & 0.676 & $+0.020$ [-0.014, +0.060] & $+0.006$ [-0.019, +0.031] & $+0.030$ [+0.007, +0.052] \\
Qwen3-14B & 1255 & 0.665 & $+0.023$ [-0.006, +0.065] & $+0.008$ [-0.017, +0.028] & $+0.027$ [+0.008, +0.046] \\
Qwen3-14B + thinking & 1220 & 0.700 & $+0.018$ [-0.012, +0.057] & $+0.010$ [-0.012, +0.031] & $+0.031$ [+0.007, +0.054] \\
Qwen3-32B & 1255 & 0.728 & $+0.050$ [+0.013, +0.095] & $+0.020$ [-0.001, +0.048] & --- \\
\bottomrule
\end{tabular}
\caption{The judge-scale ladder: every judge scores the same trajectories against
the same sealed endpoint. ``Gap to \textsc{32b}'' is paired on shared units; intervals refit
every fold.}
\label{tab:judgescale}
\end{table*}
 \begin{figure*}[t]
\centering
\includegraphics[width=\textwidth]{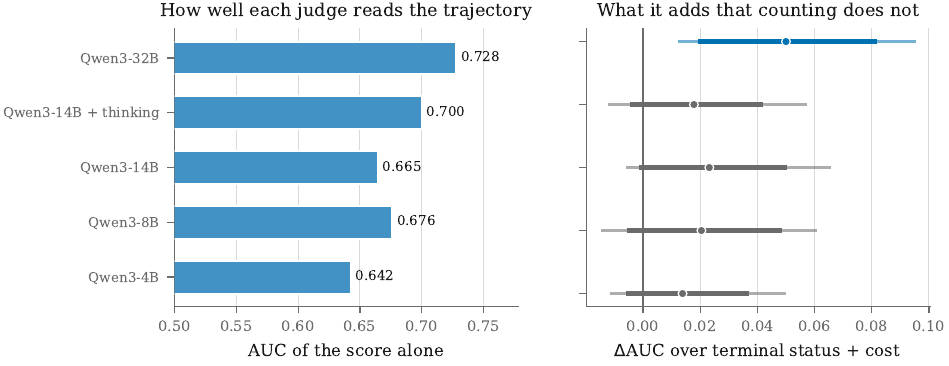}
\caption{The judge-scale ladder. Left: standalone discrimination. Right:
increment over terminal status and cost (bold: fixed-prediction interval; behind: refitted).}
\label{fig:judgescale}
\end{figure*}
\begin{table*}[t]
\centering\small
\begin{tabular}{llll}
\toprule
Baseline already contains & judge's own 0--100 & mean of five dims & five dims, free weights \\
\midrule
terminal status & $+0.099$ [+0.055, +0.145] & $+0.152$ [+0.094, +0.208] & $+0.179$ [+0.120, +0.235] \\
\quad + cost & $+0.030$ [+0.007, +0.055] & $+0.050$ [+0.020, +0.081] & $+0.055$ [+0.019, +0.089] \\
\quad + behaviour counters & $+0.012$ [-0.004, +0.029] & $+0.020$ [+0.002, +0.038] & $+0.019$ [-0.001, +0.038] \\
\bottomrule
\end{tabular}
\caption{$\Delta$AUC from adding the process score, by baseline (rows) and rubric
aggregation (columns); 1255 trajectories, 208 endpoint-positive.}
\label{tab:grid}
\end{table*}
 
\subsection{What the agents produced}

Across \Nunits{} trajectories the terminal evaluator recorded
\NtermExecutableFailing{} artifacts that run and fail, \NtermExecutablePassing{}
that run without failing, \NtermNotExecutable{} that do not run and
\NtermNoArtifact{} that submitted nothing; the sealed endpoint holds for
\ErateEThree{} (\NposEThree{} artifacts).

The endpoint is not a relabelling of the terminal status. Every $E_3$ artifact
lies inside the \textsc{executable\_failing} stratum, but only
\ErateGivenFailing{} of that stratum satisfies it: most tests that fail on the
buggy code fail for a reason the fix does not address. Terminal status is
necessary and far from sufficient, which is exactly the regime where a process
metric could earn its place.

\subsection{What it adds depends on the judge}

Table~\ref{tab:judgescale}, drawn in Figure~\ref{fig:judgescale}, is the headline: every
judge scores the same \Nunits{} trajectories against the same sealed endpoint under
the same firewall, so the judge is the only thing varying. The reasoning judge is
the one exception in coverage, returning a parsable rubric on \ThinkN{} of them;
its row carries that $n$.

The four smaller judges behave alike: at most \SmallJudgeMaxAlone{} AUC on their
own, and against terminal status plus cost the best is worth \FourteenBDelta{}
(\FourteenBCI{}), an interval covering zero. Reasoning does not change this.
Enabling thinking on the 14B raises its standalone discrimination to
\ThinkAlone{} -- a real improvement in reading the trajectory -- and leaves the
increment at \ThinkDelta{} (\ThinkCI{}). It makes a better judge, not a more
incrementally informative one.

The \BigJudgeName{} judge breaks the pattern: \BigJudgeAlone{} alone,
\BigJudgeDelta{} (\BigJudgeCI{}) over terminal status and cost, an interval that
\BigJudgeRefitExcl{} zero when every fold is refitted inside the bootstrap
(\BigJudgeRefitCI{}), and \BigJudgeBZeroDelta{} (\BigJudgeBZeroCI{}) against
terminal status alone.

That one interval clears zero and the others do not is not itself evidence that
the judges differ, so we estimate the difference. Both augmented models are built
on the same baseline predictions, so the gap between two judges' increments is the
paired contrast between their augmented models on the units carrying both scores.
The \BigJudgeName{} judge beats each of the other \LadderGapN{} by
\LadderGapMin{} (\LadderGapMinCI{}) to \LadderGapMax{} (\LadderGapMaxCI{}), and
\LadderGapAllExcl{} of those intervals exclude zero (largest
$p=\LadderGapPMax{}$). The comparison that matters most is against the reasoning
judge, which reads trajectories better than the plain 14B and still gives up
\LadderGapThinkDelta{} (\LadderGapThinkCI{}): capacity, not deliberation, is what
the increment tracks. Process information exists that is in neither the outcome nor
the cost, and reading it takes a judge of a certain size.

Most of even that is recoverable without a model: adding eight behavioural counters (Table~\ref{tab:main} in Appendix~\ref{app:sensitivity})
to the baseline reduces the largest judge's contribution to \BigJudgeBTwoDelta{}
(\BigJudgeBTwoCI{}; refitting every fold gives \BigJudgeBTwoRefitCI{}).

Table~\ref{tab:grid} varies the aggregation as
well as the baseline, and the ordering there matches the dimensions: separating the
rubric's dimensions recovers more than collapsing them, and the judge's own
aggregate is worst of the three.

\subsection{Incremental value is detectable}

A bound is only worth reporting if the same estimator would have found a larger
effect, so we answer by injection. A synthetic predictor matched to the process
score's own standalone discrimination (\ControlAUCalone{} AUC), estimated on the
same units, baseline, folds and bootstrap, moves AUC by \ControlDeltaAUC{} over
\ControlDraws{} draws and is detected in \ControlDetectRate{} of them. Sweeping
the injected strength, an injection reaching \MinDetectableAUC{} AUC alone
produces \MinDetectableDelta{} and is detected in every draw. The design is
sensitive at and above \MinDetectableDelta{}, which is above every smaller judge's
point estimate: their individual increments are bounds rather than zeros, and the
claim that they differ from the \BigJudgeName{} judge rests on the paired gap
above, which does not depend on this floor. A permutation null over whole tasks
puts the observed contrast at $p=\PermPval{}$ (Figure~\ref{fig:controls};
Appendix~\ref{app:sensitivity} explains why the test and the interval answer
different questions).

\subsection{Robustness}
\begin{table}[tbp]
\centering\small
\setlength{\tabcolsep}{2pt}
\begin{tabular}{@{}lrr@{}}
\toprule
Endpoint & $n^{+}$ & \% \\
\midrule
$E_1$ fail-to-pass & 213 & 17.0 \\
$E_2$ \quad + replay-stable & 213 & 17.0 \\
$E_3$ \quad + specific (primary) & 208 & 16.6 \\
\bottomrule
\end{tabular}

\medskip
\begin{tabular}{@{}lcc@{}}
\toprule
Endpoint & $\Delta$AUC & OR (median split) \\
\midrule
$E_1$ & $+0.053$ [+0.023, +0.084] & 4.58 [3.33, 6.29] \\
$E_2$ & $+0.053$ [+0.023, +0.084] & 4.58 [3.33, 6.29] \\
$E_3$ & $+0.050$ [+0.020, +0.081] & 4.36 [3.17, 6.00] \\
\bottomrule
\end{tabular}
\caption{The three nested endpoint definitions. Each successive requirement
removes artifacts, and none of them changes the verdict on incremental value.}
\label{tab:endpoint}
\end{table}
 
One confound deserves a check: wall time sits in the cost baseline and the campaign
ran on two accelerators, so it partly encodes which host ran an episode. Deleting
it leaves the baseline at \CostNoWallBaseAUC{} AUC against \CostFullBaseAUC{} and
the contrast at \CostNoWallDelta{} (\CostNoWallCI{}); token counts alone give
\CostTokensOnlyBaseAUC{} and \CostTokensOnlyDelta{} (\CostTokensOnlyCI{}). The
baseline knows how much the agent wrote, not how long the machine took.

Otherwise the verdict does not depend on the choices the protocol had to fix: the
three nested endpoints agree (Table~\ref{tab:endpoint}), as do regularisation
constants spanning two orders of magnitude and both aggregates
(Table~\ref{tab:sensitivity} in Appendix~\ref{app:sensitivity}), and a linear term is not hiding curvature -- the
score with its square and its interactions with terminal status gives
\FlexDeltaAUC{} (\FlexCIAUC{}). Leave-one-repository-out is noisy by
construction: \LoroNpositive{} of \LoroNfolds{} held-out repositories show a
positive point estimate, but two contribute no endpoint-positive artifact at all
and a third contributes one, so the pooled interval and not the per-fold spread is
what bounds the effect (Figure~\ref{fig:loro}).

\subsection{Across actors, judges and blocks}

Table~\ref{tab:generality} and Figure~\ref{fig:generality}, both in Appendix~\ref{app:extra}, repeat the contrast on each slice, and the pooled
estimate is not homogeneous. The contrast is \ActorFourBDelta{}
(\ActorFourBCI{}) on the 4B actor's trajectories, \ActorEightBDelta{}
(\ActorEightBCI{}) on the 8B's and \ActorFourteenBDelta{} (\ActorFourteenBCI{}) on
the 14B's -- a wider spread than the judge ladder itself.

The baselines explain the spread and point the same way as everything else here.
Terminal status plus cost reaches \ActorFourBBase{} AUC on the 4B actor's
trajectories against \ActorEightBBase{} and \ActorFourteenBBase{} on the other
two: the weaker actor's visible outcome says least about the sealed one, and that
is exactly where the process score is worth most. It is the stratum result
(\S\ref{sec:analysis}) in a second form. The reading it forbids is that a process
metric is worth \BigJudgeDelta{} on any trajectory pool; the reading it supports is
that process value appears where the outcome is uninformative, and the pooled
number is an average over pools that differ in how uninformative they are.

The pilot and confirmatory blocks separate too. On the \BlockPilotN{} trajectories
that existed when the two post-pilot specification choices were made the contrast is
\BlockPilotDelta{} (\BlockPilotCI{}); on the \BlockConfirmN{} that did not it is
\BlockConfirmDelta{} (\BlockConfirmCI{}). The confirmatory estimate is the larger
of the two, so pooling if anything understates the headline \BigJudgeDelta{}. The
confirmatory block is not an independent replication: the 4B actor's episodes are
entirely confirmatory and are the slice with the largest contrast, so block and
actor composition are confounded and the pooled number is the summary the design
supports. Appendix~\ref{app:rebuttal} repeats the ladder with a second judge
family, a non-Qwen actor and three published trajectory metrics, and tests the
score under decision rules.

 \section{Where Apparent Value Comes From}
\label{sec:analysis}

A process metric's value is a property of the judge, and so is what inflates it.

\subsection{Outcome disclosure inflates value}
\begin{table*}[t]
\centering\small
\setlength{\tabcolsep}{4pt}
\begin{tabular}{lrrll}
\toprule
Evidence condition & $n$ & AUC alone & $\Delta$AUC vs.\ \textsc{firewalled} (paired)
& $\Delta$AUC over terminal+cost \\
\midrule
\textsc{firewalled} (trajectory only) & 1255 & 0.728 & --- & $+0.050$ [+0.020, +0.081] \\
\textsc{reference} (+ sealed gold patch) & 1255 & 0.744 & $+0.016$ [-0.012, +0.043], $p=0.26$ & $+0.052$ [+0.024, +0.082] \\
\textsc{terminal} (+ terminal verdict) & 1255 & 0.831 & $+0.103$ [+0.061, +0.149], $p=0.00$ & $+0.056$ [+0.027, +0.087] \\
\textsc{redacted} (tool outputs withheld) & 1255 & 0.718 & $-0.010$ [-0.070, +0.050], $p=0.73$ & $+0.036$ [+0.013, +0.061] \\
\bottomrule
\end{tabular}
\caption{Four evidence conditions under one judge (Qwen3-32B) and rubric; rows
differ only in the evidence block shown. Intervals hold the out-of-fold predictions fixed.}
\label{tab:evidence}
\end{table*}
 \begin{figure*}[t]
\centering
\includegraphics[width=\textwidth]{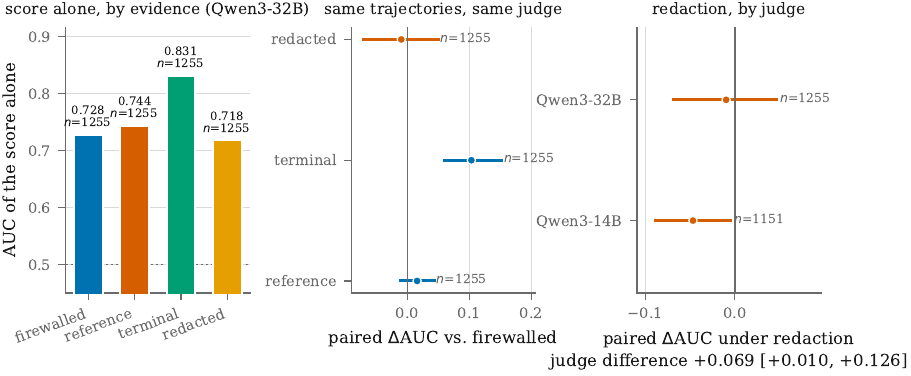}
\caption{Evidence conditions under the primary judge. Left: standalone
discrimination. Centre: paired contrast against \textsc{firewalled}. Right: redaction
applied to two judges, with their paired difference.}
\label{fig:evidence}
\end{figure*}

Table~\ref{tab:evidence} runs one judge
(\EvidenceJudge{}), one rubric and one set of trajectories under four evidence
conditions, so each row differs from the next only in what the judge was shown.
The decisive row is \textsc{terminal}, where the judge is
told what the terminal evaluator found. Its standalone discrimination rises to
\CondTerminalAUCalone{}, a paired gain of \TermLeakDeltaAUC{}
(\TermLeakCIAUC{}) over the firewalled judge on the same units.

That alone would be unsurprising: a judge told the outcome can repeat it. The
diagnostic is what follows. The outcome-disclosed score shows
\CondTerminalDelta{} (\CondTerminalCI{}) of \emph{incremental} value over a
baseline that already contains the terminal status -- and genuine information
cannot appear twice. What it adds is a re-encoding of the verdict the model can
exploit, indistinguishable in an output table from process insight
(Figure~\ref{fig:evidence}).

The claim rests on the standalone inflation, which needs no incremental model: the
same rubric on the same trajectories discriminates \BigDiscDelta{} better
(\BigDiscCI{}) when the judge is told the verdict, and
Appendix~\ref{app:sensitivity} shows the incremental part does not survive a
baseline that already holds the firewalled score and its interactions with the
verdict. A harness that discloses the outcome is not measuring the same quantity as
one that does not.

\subsection{Reference disclosure is harmless}

We expected the reference to be the dangerous leak, since a judge holding the gold
patch can check the artifact against it. It is not, for this judge: showing the
sealed patch moves standalone discrimination by \LeakDeltaAUC{} (\LeakCIAUC{},
$p=\LeakPval{}$), an interval that \LeakExcl{} zero, with the two scores
correlated at \LeakCorr{}. Reference access is not what inflates a process metric;
outcome access is.

\subsection{Redaction costs the judge nothing}

Under \textsc{redacted} every raw tool output is replaced by a placeholder, so only
the agent's own account of its work survives. We compare judges on one set of
units, because the 14B returns a parsable rubric on fewer trajectories: on the
\RedactInterN{} trajectories carrying all four scores the deletion moves the
primary judge \BigRedactMatchedDelta{} (\BigRedactMatchedCI{}), an interval
that covers zero, while the 14B moves \MidRedactMatchedDelta{}
(\MidRedactMatchedCI{}, uncorrected $p=\MidRedactP{}$), an interval that does not:
the smaller judge loses discrimination when the environment is withheld. The
paired difference between the two effects is \RedactInterDelta{}
(\RedactInterCI{}, $p=\RedactInterP{}$), which \RedactInterExcl{} zero.
The ladder's conclusion extends to the firewall: how much a judge needs to see the
environment is a property of the judge.

Surviving redaction is not a licence to trust the score: a judgement that holds up
when the environment is deleted may be reading the agent's self-report, which
redaction leaves intact. What it establishes is narrower. For the primary judge the
increment in \S\ref{sec:results} is not an artifact of raw evidence a deployment
might withhold, and the manipulation costs at most the upper end of
\BigRedactMatchedCI{}.

\subsection{The judge's aggregate loses signal}
\begin{figure}[t]
\centering
\includegraphics[width=\columnwidth]{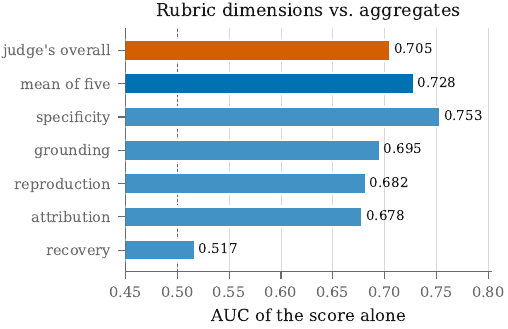}
\caption{Standalone discrimination of each rubric dimension and of the two
aggregates.}
\label{fig:dims}
\end{figure}

The five rubric dimensions are not equally informative: \BestDimName{} reaches
\BestDimAUC{} on its own and \WorstDimName{}, the one dimension not tied to the
endpoint's definition, only \WorstDimAUC{}, and their
unweighted mean reaches \MeanFiveAUCalone{}. The judge's own $0$--$100$ overall
field, produced in the same call from the same evidence, reaches only
\OverallAUCalone{}, which is \OverallVsMeanGap{} AUC worse than averaging the
numbers it had just written (Figure~\ref{fig:dims}).

This is Table~\ref{tab:grid} seen from the other side: storing the five dimensions
recovers \GridBOneFree{} over the operational baseline where the judge's own
aggregate recovers \GridBOneOverall{}. One dimension carries this alone:
\RefitBestDimName{} is worth \RefitBestDimDelta{} over terminal status and cost
with a refitted interval of \RefitBestDimCI{} that \RefitBestDimExcl{} zero
(uncorrected $p=\RefitBestDimP{}$). It is the only quantity here surviving the
refitted interval, and one of five dimensions examined, so it is close to what five
tests yield by chance: the most promising thing in the data, not a result.

\subsection{Where the value actually is}

Outside the \textsc{executable\_failing} stratum the sealed endpoint is zero by
construction, so those \StratOutN{} trajectories dilute the contrast without
contributing signal. Inside it the endpoint is genuinely uncertain: \StratN{}
artifacts all run and all fail, and \StratNpos{} satisfy the endpoint.

There the process score is worth \StratDeltaAUC{} AUC over cost, an interval that
\StratExcl{} zero with out-of-fold predictions held fixed (\StratCIAUC{}) and
\StratRefitExcl{} zero when every fold is refitted (\StratRefitCI{}). Split at the
median score, process-supported artifacts satisfy the endpoint at
\StratRateSupported{} against \StratRateDeficient{}
(Figure~\ref{fig:contingency}). This is the largest point estimate in the paper.
The stratum is nested in the pooled sample, not disjoint from it, so we do not test
the two against each other; what it shows on its own is an effect surviving the
refitting interval in the regime where the visible outcome settles least.
Appendix~\ref{app:extra} adds a matched within-task comparison and
Appendix~\ref{app:stability} shows the 14B's null is not decoding noise.
 \section{Conclusion}
\label{sec:conclusion}

Process value is gated on the judge. Against a sealed consequence and an
operational baseline, four judges from \SmallJudgeRange{} add nothing an interval
separates from zero while a \BigJudgeName{} judge adds \BigJudgeDelta{}
(\BigJudgeCI{}), and the paired gap between them is estimated rather than inferred
from overlapping intervals. What the judge may see moves it further still: the
verdict raises its discrimination by \BigDiscDelta{}, the reference and redaction
do not. A process metric reported without its judge's scale and evidence access
does not identify a quantity. 

\section*{Limitations}
Every trajectory here is repository-level regression-test generation on Python
projects with English issue text, and every endpoint is a replay of the produced
test. The design transfers to web agents, dialogue, or endpoints such as
incident rate, but the numbers do not. The main analysis uses AWQ-quantised
Qwen3 actors and judges; Appendix~\ref{app:rebuttal} adds a second judge family
and a non-Qwen actor, which is evidence from two families rather than a general
law of scale. Endpoint positives are concentrated:
\TopRepoPosShare{} come from two repositories and two contribute none, so
cross-repository transfer rests mainly on two folds. The judge-scale threshold is bracketed,
not located: we observe nothing at or below 14B and a clear effect at 32B, and
cannot say where the transition lies, whether it is scale as such rather than
some correlate of it, or what a still larger judge would find. Scoring is post
hoc by construction; Appendix~\ref{app:rebuttal} gives first evidence on early
prefixes and retries, but using the score to route or restart a live agent is
future work. Finally, our primary score is a prompted frozen rubric of our own
rather than a trained process reward model or human process annotation; three
published trajectory metrics are run through the same protocol in
Appendix~\ref{app:rebuttal}.

\clearpage
\appendix
\setcounter{topnumber}{4}\setcounter{dbltopnumber}{4}\setcounter{totalnumber}{8}
\renewcommand{\topfraction}{0.95}\renewcommand{\dbltopfraction}{0.95}
\renewcommand{\textfraction}{0.05}\renewcommand{\floatpagefraction}{0.8}\renewcommand{\dblfloatpagefraction}{0.8}
\section{Substrate detail}
\label{app:substrate}

\begin{figure}[tbp]
\centering
\includegraphics[width=\columnwidth]{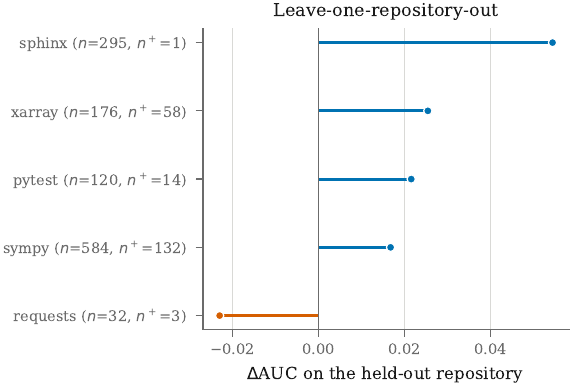}
\caption{Leave-one-repository-out contrasts. Only five of the seven folds have
an AUC: \texttt{pylint} and \texttt{seaborn} contribute no endpoint-positive
artifact. Two repositories carry
\TopRepoShare{} of the pool, so the small folds are noisy; \LoroNpositive{} of
\LoroNfolds{} folds have a positive point estimate.}
\label{fig:loro}
\end{figure}

Table~\ref{tab:repos} gives the per-repository task counts, trajectory counts and
primary-endpoint rates. Two repositories supply \TopRepoShare{} of the tasks, and
the endpoint positives are more concentrated still: \TopRepoPosShare{} of the
\NposEThree{} positive artifacts come from two repositories, two repositories
contribute none and a third contributes one. This is why every
leave-one-repository-out fold is reported individually in Figure~\ref{fig:loro}
rather than averaged, and why the per-fold estimates should be read as a
dispersion check rather than as seven independent replications.

\begin{table}[t]
\centering\small
\begin{tabular}{lrrrr}
\toprule
Repository & Tasks & Traj. & $E_3$ & \% \\
\midrule
\texttt{sympy} & 73 & 584 & 132 & 22.6 \\
\texttt{sphinx} & 38 & 295 & 1 & 0.3 \\
\texttt{xarray} & 22 & 176 & 58 & 33.0 \\
\texttt{pytest} & 15 & 120 & 14 & 11.7 \\
\texttt{requests} & 4 & 32 & 3 & 9.4 \\
\texttt{pylint} & 4 & 32 & 0 & 0.0 \\
\texttt{seaborn} & 2 & 16 & 0 & 0.0 \\
\bottomrule
\end{tabular}
\caption{Task pool composition and primary-endpoint counts per repository.}
\label{tab:repos}
\end{table}
 
Environments are per-repository virtual environments pinned to the version the
task's base commit requires. Every task in the pool was previously validated to
have a gold patch that applies cleanly and a reference test that fails before the
patch and passes after, so a failure observed in this study belongs to the
generated artifact.

\section{Serving and execution settings}
\label{app:settings}

\paragraph{Serving.} vLLM \citep{kwon2023vllm} with \texttt{--quantization
awq\_marlin} and \texttt{--max-model-len 14336}, one model resident at a time on
one card. Concurrency is a throughput setting and differs by host
(\texttt{--max-num-seqs} $24$ on the A40, $64$ on the H100); it cannot affect a
verdict, because every judge decodes at temperature $0$ with a fixed seed and
every endpoint verdict comes from executing code rather than from a sampled
token. Actors decode at temperature $0.7$,
top-$p$ $0.8$, at most $900$ new tokens per turn, with a seed derived from the
task id and the rollout index. Scorers decode at temperature $0$, top-$p$ $1.0$,
at most $400$ new tokens, except the reasoning judge, which is given $1400$ to
leave room for its chain of thought. Thinking mode is set explicitly per judge
through \texttt{chat\_template\_kwargs} and is off everywhere except that judge.
The \texttt{Qwen3-32B-AWQ} judge is served with the same flags at a larger
\texttt{--gpu-memory-utilization}; no other setting differs between judges.

\paragraph{Execution.} \texttt{pytest} is invoked as \texttt{-rA -q --tb=no -p
no:cacheprovider -p no:randomly --continue-on-collection-errors --timeout=20
--timeout-method=signal}. The signal timeout method matters: a thread-based
watchdog kills the whole session without a summary, which is indistinguishable
from a process that never started. Each execution runs in a fresh
\texttt{git worktree} hard-reset to the base commit and cleaned of untracked
files, inside \texttt{unshare -rn} with loopback brought up, with
\texttt{PYTHONHASHSEED=0}, \texttt{PYTHONDONTWRITEBYTECODE=1},
\texttt{MPLBACKEND=Agg} and a fixed locale. A run that produces no \texttt{pytest}
output at all is retried twice and then recorded as an executor failure rather
than as a verdict.

\section{Prompts}
\label{app:prompts}

\paragraph{Actor.} The system prompt states the task (write a regression test that
fails because of the described bug and would pass once it is fixed), that the fix
is not in the checkout and cannot be seen, and that the submitted file is what is
judged. The user message carries the repository name, the target path, the issue
text truncated to $6000$ characters, and the tool documentation. Actions are a
single \texttt{<action name="..."{}>} block with plain-text fields; a malformed
reply is returned to the actor as an observation and counted, never silently
repaired. The action budget is $14$; if the actor never submits, its last
\texttt{run\_test} file is taken as the submission.

\paragraph{Scorer.} The rubric asks for five $1$--$5$ dimensions (grounding,
reproduction, attribution, specificity, recovery) and one $0$--$100$ overall score,
returned as a single JSON object. The judge is told to rate the quality of the work
and is never asked to predict an outcome. The rendered trajectory contains the
issue, every turn's action and observation, and the submitted file, capped at
\TrajCharCap{} characters with the middle elided when longer, which affects
\TruncatedPct{} of trajectories; the cap is identical across evidence
conditions, so it cannot explain a between-condition contrast.

\paragraph{Evidence conditions.} The four conditions prepend one block to the
identical rendering: nothing (\textsc{firewalled}); the sealed gold patch
(\textsc{reference}); a note that tool outputs are withheld, with every observation
replaced by a placeholder (\textsc{redacted}); or the terminal evaluator's verdict
in words (\textsc{terminal}). The exact strings and their hashes are in the
released protocol.

\section{Sensitivity and additional results}
\label{app:sensitivity}
\begin{table*}[t]
\centering\small
\begin{tabular}{lrrrlr}
\toprule
Baseline & $n$ & AUC base & AUC + score & $\Delta$AUC [95\% CI] & $\Delta$log loss \\
\midrule
Terminal status & 1255 & 0.668 & 0.820 & $+0.152$ [+0.094, +0.208] & $-0.0432$ \\
\quad + cost & 1255 & 0.826 & 0.876 & $+0.050$ [+0.020, +0.081] & $-0.0384$ \\
\quad + cost + behaviour counters & 1255 & 0.827 & 0.847 & $+0.020$ [+0.002, +0.038] & $-0.0284$ \\
By task instead of repository & 1255 & 0.852 & 0.883 & $+0.031$ [+0.002, +0.060] & $-0.0250$ \\
Endpoint-conditioned stratum & 722 & 0.644 & 0.746 & $+0.103$ [+0.044, +0.169] & $-0.0679$ \\
\bottomrule
\end{tabular}
\caption{Incremental value of the process score over increasingly strong
baselines, out of fold; positive $\Delta$AUC and negative $\Delta$log loss favour the score.
The primary row is \emph{+ cost}.}
\label{tab:main}
\end{table*}
 \begin{table}[t]
\centering\small
\begin{tabular}{ll}
\toprule
Setting & $\Delta$AUC over terminal+cost \\
\midrule
$C=0.1$ & $+0.048$ [+0.014, +0.082] \\
$C=1.0$ & $+0.050$ [+0.020, +0.081] \\
$C=10.0$ & $+0.050$ [+0.022, +0.080] \\
\midrule
mean of five dimensions & $+0.050$ [+0.020, +0.081] \\
judge's own overall & $+0.030$ [+0.007, +0.055] \\
\bottomrule
\end{tabular}
\caption{The headline contrast under settings the protocol fixed in advance.}
\label{tab:sensitivity}
\end{table}
 \begin{table*}[t]
\centering\small
\begin{tabular}{lrrl}
\toprule
Score & Mean & AUC alone & $\Delta$AUC over terminal+cost \\
\midrule
specificity & 3.00 & 0.753 & $+0.060$ [+0.027, +0.092] \\
grounding & 2.91 & 0.695 & $+0.044$ [+0.014, +0.073] \\
reproduction & 2.29 & 0.682 & $+0.036$ [+0.008, +0.067] \\
attribution & 2.10 & 0.678 & $+0.037$ [+0.007, +0.069] \\
recovery & 1.26 & 0.517 & $-0.002$ [-0.013, +0.010] \\
\midrule
mean of five dimensions & --- & 0.728 & $+0.050$ [+0.020, +0.081] \\
judge's own overall & --- & 0.705 & $+0.030$ [+0.007, +0.055] \\
five dimensions, freely weighted & --- & --- & $+0.055$ [+0.019, +0.089] \\
\bottomrule
\end{tabular}
\caption{Rubric dimensions under the firewalled condition: the judge's own
$0$--$100$ aggregate discriminates worse than the mean of its five dimensions. A dash marks
a quantity the row does not have.}
\label{tab:dims}
\end{table*}
 
\paragraph{Multiplicity.} The paper reports many contrasts and corrects none of
them in the main text; Appendix~\ref{app:rebuttal} reports Holm and
Benjamini--Hochberg adjustments. There is a single
frozen-protocol primary contrast: the process score's incremental AUC over terminal
status plus cost, out of fold, grouped by repository, on the primary endpoint. The
judge ladder and the evidence conditions are pre-declared factors of that same
contrast rather than a search over specifications. Everything else -- the five
rubric dimensions, the endpoint variants, the actor and block slices, the
regularisation sweep -- is secondary and uncorrected: the dimension
result in \S\ref{sec:analysis} is the only one of its family of five to clear an
interval, which makes it the most promising
thing in the data rather than a finding. Two quantities are also reported under
two different bootstraps by design, the fixed-prediction interval and the refitting
one; the refitting interval governs, and every table says which it carries.

\paragraph{Is the contamination incremental or a re-encoding?} A categorical term
for the verdict cannot express a genuine process-by-verdict interaction, so part of
the outcome-disclosed score's apparent incremental value could be structure the
baseline cannot represent. Against a baseline already holding the firewalled score
and its interactions with the verdict, the refitted interval is \RefitContamCI{},
which \RefitContamExcl{} zero. The contamination claim in \S\ref{sec:analysis}
therefore rests on the standalone inflation and not on this quantity.

\paragraph{Interval versus test.} The cluster bootstrap interval on the
frozen-protocol contrast \MainExcl{} zero, while the permutation test against the
same baseline rejects exact independence at $p=\PermPval{}$. The permutation null,
built by permuting the score across whole tasks and rerunning the out-of-fold
procedure, has a 95th percentile of \PermNullPninetyfive{}. The two answer different nulls. The permutation null holds
the baseline's out-of-fold predictions fixed and destroys only the score's
association with the endpoint, so its spread reflects nothing but that one
association: it is a sharp test of exact independence. The bootstrap resamples
tasks and so propagates the sampling variability of the endpoint, of both models'
predictions and of the metric itself, which is what an effect-size statement
requires. A small, consistent, non-zero effect produces exactly this pair of
answers. On this aggregate the interval bounds the effect above by
\MainUpperBound{}; the test rules out exact independence, not smallness.

Table~\ref{tab:main} gives the same contrast as the middle column of
Table~\ref{tab:grid} with the other metrics attached, plus the two secondary
groupings and the endpoint-conditioned stratum.

The regularisation constant was fixed at $C=1$ in the protocol. The frozen-protocol
contrast is re-estimated at $C \in \{0.1, 1, 10\}$ in Table~\ref{tab:sensitivity};
the point estimate is stable to the third decimal and the interval sits against
zero at every setting, so the reading does not turn on this choice, but neither is
it comfortably away from zero at any of them. Per-dimension incremental values and
the alternative aggregators are in Table~\ref{tab:dims}, and the same contrast as
a function of baseline and aggregation is in Table~\ref{tab:grid}.

\section{Additional figures and slices}
\label{app:extra}
\begin{table}[t]
\centering\scriptsize
\begin{tabular}{lrl}
\toprule
Slice & $n$ & $\Delta$AUC over terminal+cost \\
\midrule
actor Qwen3-14B & 316 & $+0.012$ [-0.003, +0.028] \\
actor Qwen3-4B & 315 & $+0.110$ [+0.039, +0.183] \\
actor Qwen3-8B & 624 & $+0.020$ [-0.009, +0.050] \\
judge Qwen3-14B & 1255 & $+0.023$ [-0.000, +0.049] \\
judge Qwen3-14B (thinking) & 1220 & $+0.018$ [-0.003, +0.041] \\
judge Qwen3-4B & 1255 & $+0.014$ [-0.005, +0.036] \\
judge Qwen3-8B & 1255 & $+0.020$ [-0.005, +0.048] \\
pilot rollouts & 312 & $+0.031$ [-0.004, +0.066] \\
confirmatory rollouts & 943 & $+0.061$ [+0.029, +0.095] \\
\bottomrule
\end{tabular}
\caption{The incremental contrast under each actor, each judge scale, and the
pilot versus confirmatory rollout blocks.}
\label{tab:generality}
\end{table}
 
\paragraph{Matched within-task pairs.} The within-task comparison removes task difficulty too. Among \PairN{} pairs on the
same task by the same actor with the same terminal status and opposite endpoints,
the score ranks the endpoint-satisfying member higher \PairConcordant{} times,
lower \PairDiscordant{} times, and gives the two members an identical score in the
remaining \PairTies{}. On the \PairDecided{} pairs it decides, the score is right
\PairRate{} of the time (\PairCI{}, sign test $p=\PairP{}$), which is
\PairVerdict{}. The proportion, the interval and the test all refer to those
\PairDecided{} pairs; counting the ties as half-successes instead gives
\PairRateTiesSplit{}. The pairs share repository, issue, model and visible outcome and
differ only in the sealed consequence, which makes this the cleanest comparison
available and, at \PairN{} pairs, the thinnest.

\begin{figure}[tbp]
\centering
\includegraphics[width=\columnwidth]{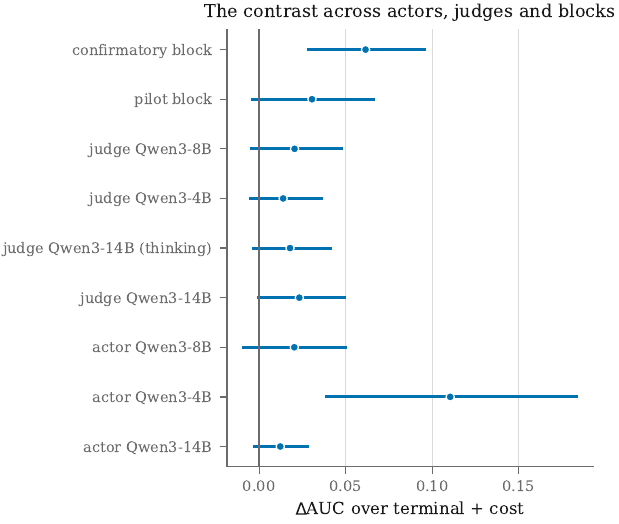}
\caption{The incremental contrast on each slice of the data: each actor, each
judge scale, and the pilot versus confirmatory rollout blocks.}
\label{fig:generality}
\end{figure}

\begin{figure*}[tbp]
\centering
\includegraphics[width=\textwidth]{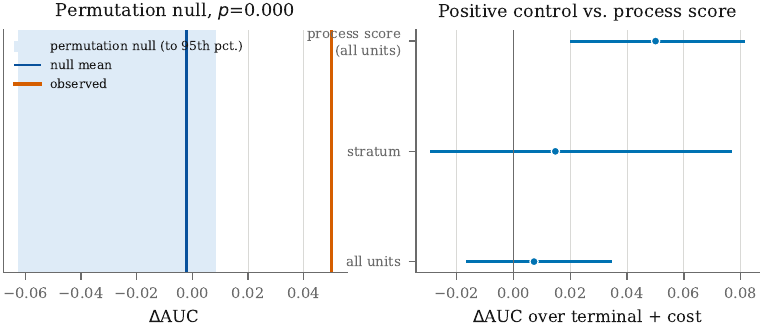}
\caption{Left: the permutation null for the incremental contrast, with the
observed value marked. Right: the synthetic matched-strength predictor and the
process score, estimated identically on the same units. The design detects
incremental value when it is present.}
\label{fig:controls}
\end{figure*}

\begin{figure}[b]
\centering
\includegraphics[width=\columnwidth]{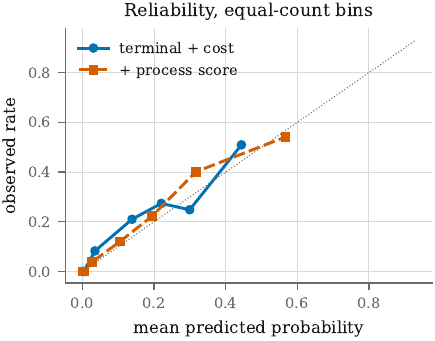}
\caption{Reliability of the two out-of-fold models, equal-count bins.}
\label{fig:reliability}
\end{figure}

Table~\ref{tab:generality} lists every slice over which the contrast is estimated,
together with the contrast on that slice. Figures~\ref{fig:reliability},
\ref{fig:loro}, \ref{fig:generality}, \ref{fig:controls},
\ref{fig:contingency}, and~\ref{fig:riskcov} give the remaining diagnostics:
calibration of the two out-of-fold models, the per-repository contrasts, the same
contrast on each slice, the permutation null beside the matched-strength
synthetic control, the endpoint rate by terminal status and median score split,
and the risk-coverage curves.

\section{A second, weaker control}
\label{app:goldoverlap}

A second control feature comes from the sealed patch: the fraction of
identifiers the gold patch touches that also
appear in the submitted artifact. It reaches \GoldOverlapAUCalone{} AUC alone and
moves the estimator by \GoldOverlapDelta{} (\GoldOverlapCI{}) -- too weak to
serve as a power statement, so the
matched-strength synthetic predictor of \S\ref{sec:results} fills that role.

\section{Judge stability and decision utility}
\label{app:stability}

A tempting explanation of the small judges' nulls is that a prompted judge is
noise. It is not. The reliability run is on the \textbf{14B} judge, decoded three
times at temperature $1$ (the primary judge decodes at temperature $0$ and
reproduces exactly, so it has no seed variance to measure at temperature $0$;
Appendix~\ref{app:rebuttal} adds a temperature-$1$ run). Across those three
seeds the firewalled score has reliability \ScorerICC{} and mean pairwise
correlation \ScorerSeedCorr{}, and averaging them -- the standard remedy for
attenuation -- gives standalone discrimination \ScorerAvgAUC{} and a contrast of
\ScorerAvgDelta{} (\ScorerAvgCI{}). Every number in this paragraph is a 14B number
and none of them bears on the \BigJudgeName{} result. What they establish is that
the 14B's null is not an artifact of decoding noise: de-noising it by averaging
does not lift it to the \BigJudgeName{} judge's \BigJudgeDelta{}. A small judge
measures something stable, and what it measures is largely already in the outcome
and the cost.

\begin{figure*}[tbp]
\begin{minipage}[t]{0.48\textwidth}
\centering
\includegraphics[width=\linewidth]{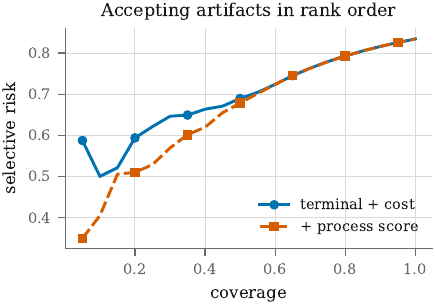}
\caption{Selective risk when artifacts are accepted in descending order of
predicted endpoint probability.}
\label{fig:riskcov}
\end{minipage}\hfill
\begin{minipage}[t]{0.48\textwidth}
\centering
\includegraphics[width=\linewidth]{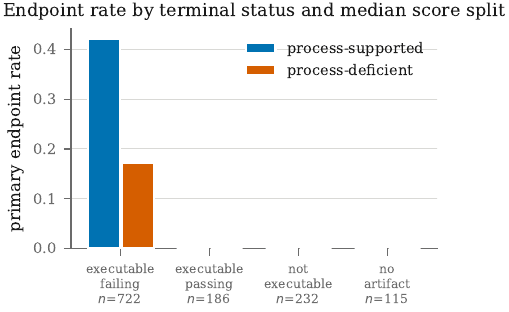}
\caption{Primary-endpoint rate by terminal status, split at the median process
score. Outside the \textsc{executable\_failing} stratum the endpoint is zero by
construction. Inside it the split separates, \StratRateSupported{} against
\StratRateDeficient{}, and the corresponding out-of-fold contrast survives
refitting (\S\ref{sec:analysis}).}
\label{fig:contingency}
\end{minipage}
\end{figure*}

The decision-utility view belongs to the primary \BigJudgeName{} judge, whose
score feeds the augmented model: accepting artifacts in descending order of
predicted endpoint probability \citep{geifman2017selective}, adding that score to
the operational baseline moves the area under the risk-coverage curve by
\AurcDelta{} (\AurcCI{}, which \AurcExcl{} zero; Figure~\ref{fig:riskcov}). It is
the decision-metric face of the incremental value in \S\ref{sec:results}, not a
property of the 14B.

\section{Endpoint composition}
\label{app:endpoint}

Table~\ref{tab:endpoint} reports what each successive requirement removes. Replay
stability removes nothing: every $E_1$ artifact reproduced its verdict exactly, so
$E_2$ and $E_1$ are the same set here and their agreement in the table is
definitional rather than evidential. The specificity check against an unrelated
patch is the requirement that binds, removing \EthreeRemoved{} artifacts that flip
under any perturbation of the source.

\section{Second judge family, decision rules}
\label{app:rebuttal}

This appendix runs the protocol of \S\ref{sec:method} on the same \Nunits{}
trajectories and sealed endpoint with judges outside the main ladder and with
three published trajectory metrics, and then asks what the primary score is good
for. $\Delta B_1$ is the increment over terminal status and cost; $\Delta B_2$
adds the eight behavioural counters. These runs are additional to the compute
reported in \S\ref{sec:setup}.

\begin{table}[h]
\centering\small
\setlength{\tabcolsep}{4pt}
\begin{tabular}{lrrr}
\toprule
Judge / metric & AUC alone & $\Delta B_1$ & $\Delta B_2$ \\
\midrule
Gemma-3-4B & 0.610 & $+0.002$ & $-0.001$ \\
Gemma-3-12B & 0.556 & $+0.003$ & $-0.001$ \\
Gemma-3-27B & 0.717 & $+0.049$ & $+0.019$ \\
Qwen3-30B-A3B & 0.671 & $+0.010$ & $+0.006$ \\
Qwen3-32B (primary) & 0.728 & $+0.050$ & $+0.020$ \\
\midrule
MCP-Bench & 0.749 & $+0.053$ & $+0.016$ \\
Agent-as-a-Judge & 0.718 & $+0.034$ & $+0.012$ \\
TRACE & 0.575 & $-0.006$ & $-0.000$ \\
\bottomrule
\end{tabular}
\caption{Judges from a second family, a sparse judge (3B active parameters) and
three published trajectory metrics, on the same trajectories and endpoint.}
\label{tab:extrajudges}
\end{table}

\paragraph{The same pattern in a second judge family.} Table~\ref{tab:extrajudges}
reproduces the ladder of Table~\ref{tab:judgescale} in a second family:
Gemma-3-27B matches the primary judge ($+0.049$ against $+0.050$), while the
smaller Gemma judges and the sparse Qwen3-30B-A3B add little, as the 4B--14B
Qwen3 judges do. The published metrics follow the same pattern: rubrics that
read what the artifact does (MCP-Bench, Agent-as-a-Judge) recover the signal and
TRACE's process composite does not. Once the counters are in the baseline every
row is at most $+0.020$, the result of \S\ref{sec:results}. On trajectories from
a non-Qwen actor (Llama-3.1-8B), the Qwen3-32B judge also reads trajectories
better than the 14B (AUC alone 0.726 against 0.710). Rerun at temperature $1$
over three seeds, the primary judge keeps its $+0.050$ with reliability $0.812$.

\begin{table}[h]
\centering\small
\begin{tabular}{lr}
\toprule
Decision rule & Score vs.\ baseline \\
\midrule
Filter, reject 50\%: recall & 0.971 vs.\ 0.933 \\
Selective ranking: $\Delta$AURC & $-0.036$ \\
First 4/14 actions vs.\ prefix & $+0.061$ \\
Best-of-$n$ (473 groups) & $+0.006$ \\
One-shot retry & $\approx 0$ \\
\bottomrule
\end{tabular}
\caption{The primary score under decision rules on the existing rollouts.}
\label{tab:decision}
\end{table}

\paragraph{What the score is for.} Table~\ref{tab:decision} shows the score helps
where the decision is whether to trust or flag a trajectory: it raises recall as
a post-hoc filter, lowers selective risk (Figure~\ref{fig:riskcov}), and adds
signal from the first four actions. Choosing among rollouts of one task and a
one-shot retry gain little. The score is a diagnostic for triage and early
warning, which is the incremental diagnostic value the main text measures.

\paragraph{Detectability and multiplicity.} The process score's correlation
with the baseline prediction is $0.136$ ($0.202$ with the counters), the
near-independent end of the detectability sweep, where a matched synthetic
control is detected in 30 of 30 draws (mean $+0.052$); at correlation $0.366$ it
is detected in 18 of 30 and at $0.613$ in none. Across the 22 secondary
contrasts, Holm correction retains 6 and Benjamini--Hochberg 12.
 
\end{document}